# Learning Transferability: A Two-Stage Reinforcement Learning Approach for Enhancing Quadruped Robots' Performance in U-Shaped Stair Climbing

Baixiao Huang[1†], Baiyu Huang[2†], Yu Hou, Ph.D.[3]

[1]Independent Researcher, 7200 Bollinger Rd, Apt 516, CA 95129. Email: baixiaohuang1@gmail.com ORCID: 0009-0000-9629-9665
[2]Independent Researcher, 16506 82nd Pl NE, Kenmore, WA, 98028. Email: baiyuhuang2@gmail.com ORCID: 0009-0000-7224-1987
[3]Western New England University, Department of Construction Management, 1215 Wilbraham Rd, Springfield, MA 01119 (corresponding author). Email: yu.hou@wne.edu ORCID: 0000-0002-9822-244X
†These authors contribute equally to this work

**ABSTRACT**

Quadruped robots are employed in various scenarios in building construction. However, autonomous stair climbing across different indoor staircases remains a major challenge for robot dogs to complete building construction tasks. In this project, we employed a two-stage end-to-end deep reinforcement learning (RL) approach to optimize a robot's performance on U-shaped stairs. The training robot-dog modality, *Unitree Go2*, was first trained to climb stairs on *Isaac Lab*'s pyramid-stair terrain, and then to climb a U-shaped indoor staircase using the learned policies. This project explores end-to-end RL methods that enable robot dogs to autonomously climb stairs. The results showed (1) the successful goal reached for robot dogs climbing U-shaped stairs with a stall penalty, and (2) the transferability from the policy trained on U-shaped stairs to deployment on straight, L-shaped, and spiral stair terrains, and transferability from other stair models to deployment on U-shaped terrain.

## INTRODUCTION

Robots have been widely used in building construction, particularly for repetitive, hazardous, labor-intensive, or precision-intensive tasks. Unmanned Ground Vehicles (UGVs) are among the most practical robot platforms in the construction robotics domain, as they can transport payloads, carry sensors, and operate in various environments. The applications include site logistics, construction progress tracking, safety inspection, and demolition. However, traditional UGVs for construction tasks are tracked or wheeled and can operate only at a single ground level until quadruped robots, also known as robot dogs, are employed. These robot dogs possess several distinct advantages that enable them to outperform other robots. Crucially, their inherent stability, derived from their four-legged gait, provides a robust solution for reliable movement in building construction environments. Furthermore, its extensibility for mounting robot arms, multiple sensors, and other attachments enables it to accommodate diverse use cases.

Despite these strengths, the use of robot dogs remains constrained by several challenges. The first challenge is to achieve a comprehensive semantic understanding and situational awareness. Recent studies have explored integrating Building Information Modeling (BIM) and Simultaneous Localization and Mapping (SLAM) with robotic perception to enhance the adaptive capabilities of robot dogs. The second challenge is the robust stair-climbing capabilities. To successfully complete construction tasks with minimal human operation, robots must be able to ascend and descend a variety of stairs. To the best of our knowledge, current robot stair-climbing still requires human intervention and exhibits unstable climbing performance.

In this project, we explored a two-stage end-to-end deep reinforcement learning (RL) framework for robot dogs to address stair-climbing challenges. We designed and trained our algorithm using the NVIDIA *Isaac Lab* platform and implemented it on the *Unitree Go2*. To test our approach, we conducted experiments on a U-shaped staircase to evaluate its effectiveness for supporting climbing tasks in complex environments. We aim to answer questions such as:

- **RQ1:** How can penalty strategies assist robots in achieving goals on a U-shaped stair?
- **RQ2:** How can different RL training strategies be mutually transferred between a U-shaped stair and other types?

The rest of the paper is organized as follows. Related work has discussed recent innovations in UGVs, including quadruped robots. The methodology introduced the RL framework used in this project. Experiments tested cases across different stair environments and stair-surface conditions and conducted a cross-evaluation of RL training strategies. The discussion emphasized both successful and unsuccessful cases and explained the reasons for the outcomes. The conclusion synthesized the achievements and future work of the RL framework for quadruped robots.

**RELATED WORK FROM UGV TO QUADRUPED ROBOTS**

UGV has been widely used to replace manual walk-throughs with autonomous data collection for site inspection and construction progress monitoring. (Asadi et al. 2020) designed an Unmanned Aerial Vehicle (UAV)-UGV team for construction site inspection. The system is capable of autonomous navigation to localize both UAVs and UGVs in the mapping of construction environments. (Zeng et al. 2024) employed mobile robots to assess construction safety, and researchers also used automated vehicles to generate as-built digital twins, such as Scan-to-BIM and SLAM for building reconstruction.

Recently, researchers have shifted from wheeled UGVs to legged UGVs, such as quadruped robot dogs, due to terrain irregularity, on-site debris, and the presence of indoor stairs. (Halder et al. 2023) built a human-robot team for construction inspection and monitoring with quadruped robots. (Naderi et al. 2025) employed a multilayer VLM-LLM pipeline to support autonomous construction-site safety inspections using robot dogs. (Chen et al. 2025) used a multi-sensor robot dog to capture the indoor environment. However, to the best of our knowledge, current research focuses primarily on quadrupedal robot motion at the same ground level. Researchers sometimes overlook that stairs in indoor building construction remain a challenge for quadrupedal robots. Training robots to robustly and quickly climb a diverse range of stair shapes should be investigated for their applications in building construction.

# METHODOLOGY

We trained a neural network policy to enable a robot dog to climb stairs, which takes proprioceptive states, environmental states represented as a height map, and climbing goals as inputs and observations. The neural network is trained by using deep RL with the on-policy algorithm *Proximal Policy Optimization* (PPO) (Schulman et al. 2017) in the *Isaac Lab* environment. As shown in Figure 1, there are two stages for robot performance training. The terrain in the first stage is *Isaac Lab's* default pyramid stairs, which gave the robot basic climbing skills. The second stage trains the policy on custom stair terrains, U-shaped stairs, to develop terrain-specific navigation skills.

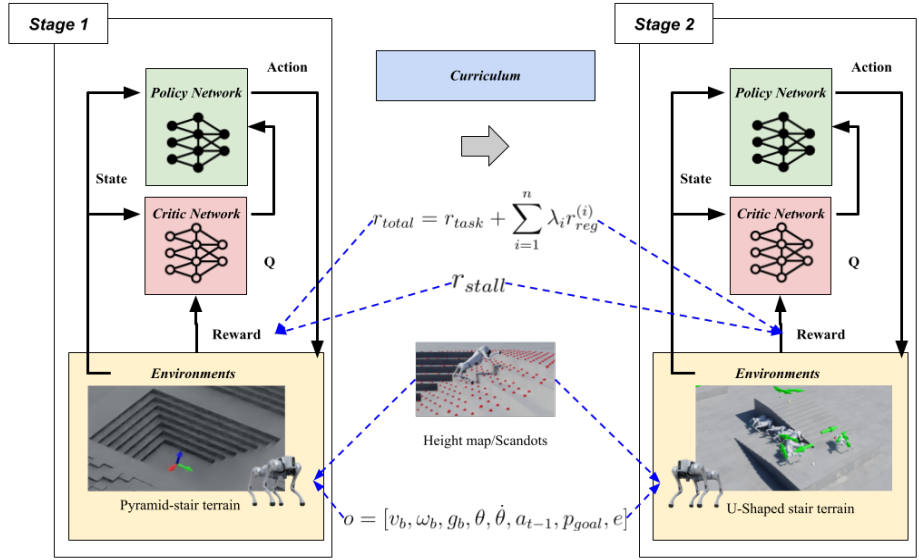

**Figure 1. A Two-Stage End-to-End RL Framework**

**Observations:** The inputs and observations fed to the neural network are $o = [v_b, \omega_b, g_b, \theta, \dot{\theta}, a_{t-1}, p_{goal}, e]$. The symbols represent the quadruped's linear velocity ($v_b \in R^3$) and angular velocity ($\omega_b \in R^3$) in the body frame, the gravity vector ($g_b \in R^3$) in the body frame, the position ($\theta \in R^{12}$) and velocity ($\dot{\theta} \in R^{12}$), the joint position command ($a_{t-1} \in R^{12}$), the goal pose ($p_{goal} = [x, y, z, \psi]^T \in R^4$), and the environmental states ($e$). These observations provide the policy with the robot's kinematic and proprioceptive states, as well as the terrain surrounding the robot.

**Network Architecture:** The policy (actor) network consists of a shallow convolutional neural network (CNN) encoder followed by a three-layer multilayer perceptron (MLP). The CNN processes a 21x21 local height-map observation and extracts a 128-dimensional latent feature vector that captures terrain geometry. This terrain embedding is concatenated with the robot proprioceptive observations described previously and then fed into the MLP. The CNN comprises convolutional and pooling layers, whereas the MLP comprises 3 hidden layers with sizes 128, 128, and 64. Our RL framework comprises a critic network with the same architecture as the policy (actor) network, and that shares the CNN encoder with it. Sharing the encoder eliminates the need to train a separate terrain-representation model and improves sample efficiency by enforcing a common perception module for control and value estimation. However, the critic maintains independent weights in its MLP because its objective is value regression rather than action generation.

**Task Rewards:** The total rewards are $r_{total} = r_{task} + \sum_{i=1}^{n} \lambda_i r_{reg}^{(i)}$, where $\lambda_i$ is the weight, and $r_{reg}^{(i)}$ represents the regularization reward and penalties. The task rewards are formulated as in our previous work (Huang et al. 2026). In stage 1 training, we used a navigation reward, computed as a function of the Euclidean distance between the robot and the goal and the reward range. We also used two navigation rewards to handle both long-distance and close-to-goal tracking. In stage 2, the navigation reward was replaced with a centering reward and a path reward, which encourage the robot to travel along the center of the stairs.

**General Penalties:** In addition to task rewards, regularization rewards penalize the quadruped for extreme behavior. Therefore, we presented notable regularization rewards, including penalties for motor power output, excessive torque, rapid action changes, reaching the physical limit, and excessive joint velocity and acceleration.

**Special Penalty - Stall Penalty:** Our previous work (Huang et al. 2026), shows promising stair-climbing results, but robot dogs hesitated to climb the stairs when the difficulty level was higher. Robots realized that the expected total rewards of climbing, with a potential penalty for falling, were lower than those of staying. To encourage robots to climb, we imposed a stall penalty, $r_{stall}$. The robot is penalized with a constant negative reward if it moves below 0.3 m/s outside the goal regions. This encourages the robot to move and explore alternative navigation strategies instead of being stuck at the starting location when presented with challenging terrain.

**Curriculum:** To assist the robot in gradually learning climbing skills, we employed a training curriculum (Lee et al. 2020). The terrain is divided into ten levels, with progressive difficulty for each level. The difficulty is determined by the stair riser height and tread width. The cumulative curriculum design can significantly improve learning speed. The rule of leveling up follows (Rudin et al. 2022)'s strategies. Given a quadruped at a specific level, it can level up to the more difficult terrain if it can reach the goal position with specified yaw. Otherwise, it gets demoted to a lower level if it fails. If it completes all levels, it is placed on a random level for continued training to prevent catastrophic forgetting of the policy. We deploy 4096 quadrupeds to train simultaneously on the terrain. Each follows the level-up rule outlined above independently.

**EXPERIMENTS**

To address the research questions we proposed, we designed the following experiments.

- **EXP1:** We trained U-shaped stairs RL strategies with general and special penalties, including the stall penalty.
- **EXP2:** We examined the transferability of RL strategies across different stair shapes, such as straight, L-shaped, U-shaped, and spiral (Huang et al. 2026).

We tested the staircase with parameters that differed from those used in training by customizing the test terrain to 6 levels, each with a different stair riser height. The policy's performance is measured using the following metrics.

- **Goal-reached success rate:** This measures the percentage of quadrupeds that successfully reach the goal location under a single terrain type and difficulty level.
- **Transferability metrics:** Transferability is defined as the ratio of a transferred model's success rate to that of a model trained specifically for the target terrain, $\text{Transferability} = \frac{S_{transfer}}{S_{target-trained}}$.

**RESULTS AND DISCUSSION**

**EXP1 U-Shaped Stair Climbing.** Figure 2 plots the success rate of reaching goals versus stair difficulty level for the U-shaped staircase. The U-shaped staircases performed well from Levels 1 to 5. However, the performance curve was monotonically decreasing, with a sharp drop at Level 6, showing that we still need to optimize our RL strategies for a more challenging staircase. We documented our robot's climbing in a YouTube video (Huang, 2026). Figure 3 presents screenshots from YouTube showing the results.

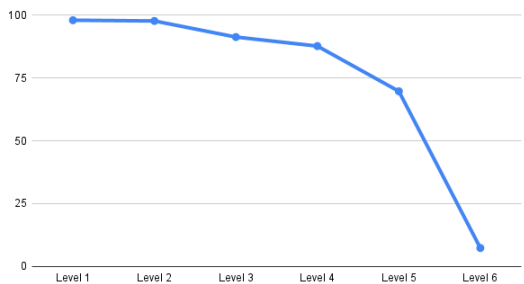
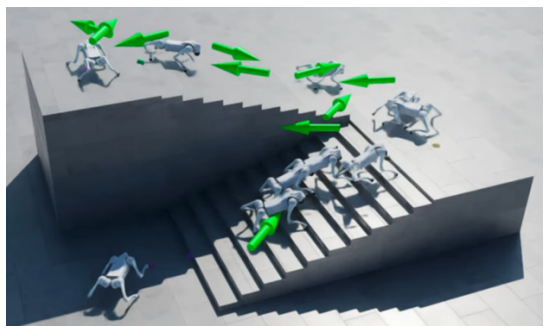

**Figure 2. Success Rate of U-shaped stair climbing**

**Figure 3. Robot Dogs Climb the U-Shaped Stairs**

To qualitatively analyze what the policy learns, we evaluated the critic network at different spatial locations along the stairs by providing a local height map and goal position. We fixed the base velocity, joint velocities, and foot-scan inputs to zero. The Critic values are evaluated on the x-y plane and visualized in Figure 4. The orange box indicates the stair entrance, the red boxes denote the two ascending runs, the blue boxes mark the landings, and the red star indicates the goal location. We observed that the value distribution aligns clearly with the stair topology. Regions near the goal, along the stair runs, and around the stair entrance show relatively higher values, indicating states from which the robot can feasibly progress toward the goal. In contrast, walls, landing edges, and geometrically constrained areas correspond to lower values, reflecting reduced long-horizon feasibility due to collision or fall risks.

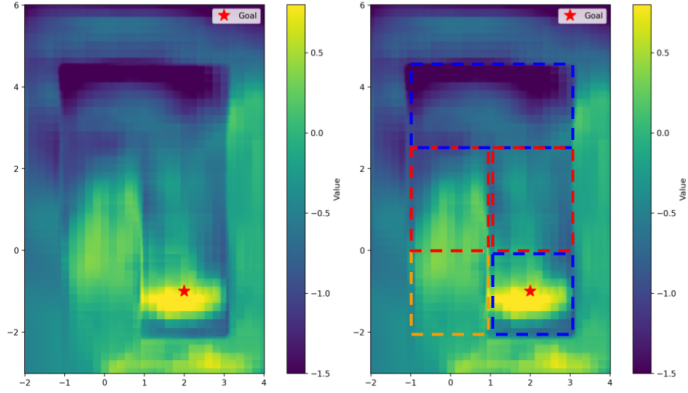

**Figure 4. Critic Values the Heat Map Evaluated on the U-Shaped Stairs**

These results suggest that the critic encodes terrain-conditioned long-horizon feasibility rather than Euclidean distance to the goal, indicating that the learned value function is sensitive to the environment's geometric structure.

**EXP2 Transferability:** We evaluated the transferability of the policy trained on U-shaped stairs by deploying it on straight, L-shaped, and spiral stair terrains at difficulty levels 3 and 4, which correspond to the medium across all six levels. Table 1 summarizes the success rates for each terrain. The total success rate is computed as the average of the Level 3 and Level 4 results, and the transferability is reported relative to the performance of the model trained specifically on the terrain of interest.

**Table 1. Testing Results for the U-Shaped Trained Model on Different Stairs**

|  | U-shaped Model at Level 3 | U-shaped Model at Level 4 | Total Success Rate | Transferability |
|---|---|---|---|---|
| U-shaped stairs | 91.3% | 87.7% | 89.5% | N/A |
| Straight stairs | 98.0% | 96.3% | 97.2% | 97.6% |
| L-shaped stairs | 75.7% | 72.1% | 73.9% | 83.6% |
| Spiral stairs | 44.7% | 25.8% | 35.3% | 50.6% |

The U-shaped trained model achieves the highest transfer performance on straight stairs, with an average success rate of 97.2% (97.6% transferability), indicating strong generalization to purely straight configurations. This is consistent with the presence of extended straight segments in the U-shaped training environment.

Transferability on L-shaped stairs is lower, with a total success rate of 73.9% (83.6% transferability), reflecting the additional geometric complexity introduced by the turn and landing. However, the model retains robust performance, suggesting that training on U-shaped stairs enables effective handling of connected segments and discrete directional changes. In contrast, performance on spiral stairs is substantially lower, with an average success rate of 35.3% (50.6% transferability). This suggests that continuously varying curvature and asymmetric step geometry in spiral staircases are not sufficiently represented in the U-shaped environment.

Overall, these results suggest that training on U-shaped stairs promotes behaviors that generalize well to terrains composed of straight segments and discrete turns, while transfer to environments with continuous curvature remains limited.

To evaluate how well models trained on different terrains transfer to the U-shaped terrain, we deployed models trained on straight, L-shaped, and spiral terrains onto the U-shaped terrain at difficulty levels 3 and 4. The total success rate is computed as the average of the Level 3 and Level 4 results. Transferability is defined as the ratio between the transferred model's success rate and that of the model trained specifically on the U-shaped terrain.

**Table 2. Transferability from Other Models to the U-Shaped Terrain**

|  | U-shaped stairs level 3 success rate | U-shaped stairs level 4 success rate | Total success rate | Transferability |
|---|---|---|---|---|
| U-shaped model | 91.3% | 87.7% | 89.5% | N/A |
| Straight model | 2.0% | 1.0% | 1.5% | 1.7% |
| L-shaped model | 0.7% | 2.0% | 1.4% | 1.5% |
| Spiral model | 1.7% | 0% | 0.9% | 1.0% |

The results in Table 2 show that models trained on non–U-shaped terrains achieve very low success rates (approximately 1–2%) when transferred to the U-shaped terrain. This indicates that the U-shaped environment presents structural and navigational challenges that are not adequately captured by the other training terrains. In contrast, as shown in the previous experiment, the model trained on the U-shaped terrain retains higher transfer performance when evaluated on Straight and L-shaped terrains. This asymmetry in transfer performance suggests that the U-shaped terrain induces more diverse or demanding locomotion–navigation behaviors, while also encompassing structural elements, such as straight stair segments and discrete turns, that generalize to simpler terrain configurations.

**CONCLUSION**

Autonomous legged robots have recently emerged as promising platforms for operation in built environments. Unlike wheeled or tracked systems, robot dogs are more flexible. Nevertheless, reliable traversal of indoor staircases remains a fundamental limitation preventing multi-floor deployment. Our two-stage end-to-end deep RL framework demonstrates promising performance in robot dogs climbing U-shaped stairs across various penalty designs. We contributed to the design of a stall penalty that encourages robot dogs to continuously explore the environment and climb stairs. On the other hand, we tested the transferability between the U-shaped terrain and other stair models. It shows that the U-shaped environment presents challenges that are not adequately captured by the other stair terrains. However, the model trained on the U-shaped terrain retains higher transfer performance when evaluated on other terrains. In the future, we plan to conduct sim-to-real experiments to validate our simulation under real-world conditions.

# REFERENCES

Asadi, K., A. Kalkunte Suresh, A. Ender, S. Gotad, S. Maniyar, S. Anand, M. Noghabaei, K. Han, E. Lobaton, and T. Wu. 2020. "An integrated UGV-UAV system for construction site data collection." Automation in Construction, 112: 103068. https://doi.org/10.1016/j.autcon.2019.103068.

Chen, Z., C. Song, B. Wang, X. Tao, X. Zhang, F. Lin, and J. C. P. Cheng. 2025. "Automated reality capture for indoor inspection using BIM and a multi-sensor quadruped robot." Automation in Construction, 170: 105930. https://doi.org/10.1016/j.autcon.2024.105930.

Halder, S., K. Afsari, E. Chiou, R. Patrick, and K. A. Hamed. 2023. "Construction inspection & monitoring with quadruped robots in future human-robot teaming: A preliminary study." Journal of Building Engineering, 65: 105814. https://doi.org/10.1016/j.jobe.2022.105814.

Huang, B., B. Huang, and Y. Hou. 2026. "Training and Simulation of Quadrupedal Robot in Adaptive Stair Climbing for Indoor Firefighting: An End-to-End Reinforcement Learning Approach." arXiv. https://doi.org/10.48550/arXiv.2602.03087

Huang, B. 2026. U-shaped stairs climbing. Online: https://youtu.be/7F6I332l7KE

Lee, J., J. Hwangbo, L. Wellhausen, V. Koltun, and M. Hutter. 2020. "Learning Quadrupedal Locomotion over Challenging Terrain." Science Robotics, 5 (47): eabc5986. https://doi.org/10.1126/scirobotics.abc5986.

Naderi, H., A. Shojaei, P. Agee, K. Afsari, and A. Akanmu. 2025. "Autonomous Construction-Site Safety Inspection Using Mobile Robots: A Multilayer VLM-LLM Pipeline." arXiv.

Rudin, N., D. Hoeller, P. Reist, and M. Hutter. 2022. "Learning to Walk in Minutes Using Massively Parallel Deep Reinforcement Learning." arXiv.

Schulman, J., F. Wolski, P. Dhariwal, A. Radford, and O. Klimov. 2017. "Proximal Policy Optimization Algorithms." arXiv.

Zeng, L., S. Guo, J. Wu, and B. Markert. 2024. "Autonomous mobile construction robots in built environment: A comprehensive review." Developments in the Built Environment, 19: 100484. https://doi.org/10.1016/j.dibe.2024.100484.